\documentclass[letterpaper, 10 pt, conference]{ieeeconf}  

\IEEEoverridecommandlockouts                              

\usepackage[T1]{fontenc}
\usepackage{hyperref}
\usepackage{graphicx}
\usepackage{amsmath}
\usepackage{amssymb}
\usepackage{booktabs}
\usepackage{tabularray}
\usepackage{multirow}
\usepackage{pifont}
\usepackage{subfigure}
\usepackage{tikz}
\usepackage{xcolor}
\graphicspath{ {./images/} }

\definecolor{car}{RGB}{100, 150, 245} 
\definecolor{bicycle}{RGB}{100, 230, 245} 
\definecolor{motorcycle}{RGB}{30, 60, 150} 
\definecolor{truck}{RGB}{80, 30, 180} 
\definecolor{other-vehicle}{RGB}{100, 80, 250} 
\definecolor{person}{RGB}{255, 30, 30} 
\definecolor{bicyclist}{RGB}{255, 40, 200} 
\definecolor{motorcyclist}{RGB}{150, 30, 90} 
\definecolor{road}{RGB}{255, 0, 255} 
\definecolor{parking}{RGB}{255, 150, 255} 
\definecolor{sidewalk}{RGB}{75, 0, 75} 
\definecolor{other-ground}{RGB}{175, 0, 75} 
\definecolor{building}{RGB}{255, 200, 0} 
\definecolor{fence}{RGB}{255, 120, 50} 
\definecolor{vegetation}{RGB}{0, 175, 0} 
\definecolor{trunk}{RGB}{135, 60, 0} 
\definecolor{terrain}{RGB}{150, 240, 80} 
\definecolor{pole}{RGB}{255, 240, 150} 
\definecolor{traffic-sign}{RGB}{255, 0, 0} 

\title{\LARGE \bf
OccupancyDETR: Using DETR for Mixed Dense-sparse 3D Occupancy Prediction$^{*}$
}

\author{Yupeng Jia$^{1}$$^{\dag}$, Jie He$^{1}$$^{\dag}$, Runze Chen$^{1}$, Fang Zhao$^{1}$ and Haiyong Luo$^{2}$$^{\ddag}$
\thanks{$^{*}$ This work was supported in part by the Strategic Priority Research Program of Chinese Academy of Sciences under Grant XDA28040500, the National Natural Science Foundation of China under Grant 62261042 and 62002026, the Key Research Projects of the Joint Research Fund for Beijing Natural Science Foundation and the Fengtai Rail Transit Frontier Research Joint Fund under Grant L221003, Beijing Natural Science Foundation under Grant 4212024 and 4222034, the Key Research and Development Project from Hebei Province under Grant 21310102D , the Fundamental Research Funds for the Central Universities under Grant 2022RC13, Yibin City Introduction of High-Level Talent Project under Grant 2022YG03 and the Open Project of the Beijing Key Laboratory of Mobile Computing and Pervasive Device, Institute of Computing Technology, Chinese Academy of Sciences (Corresponding author: Haiyong Luo)}
\thanks{$^{1}$ School of Computer Science (National Pilot Software Engineering School), Beijing University of Posts and Telecommunications, Beijing, China}%
\thanks{$^{2}$ Institute of Computing Technology, Chinese Academy of Sciences, Beijing, China}%
\thanks{$^{\dag}$ Equal contribution.}%
\thanks{$^{\ddag}$ Corresponding author, email addresses: yhluo@ict.ac.cn}%
}

\begin{document}

\maketitle
\thispagestyle{empty}
\pagestyle{empty}

\begin{abstract}

  Visual-based 3D semantic occupancy perception is a key technology for robotics, including autonomous vehicles, offering an enhanced understanding of the environment by 3D. This approach, however, typically requires more computational resources than BEV or 2D methods. We propose a novel 3D semantic occupancy perception method, OccupancyDETR, which utilizes a DETR-like object detection, a mixed dense-sparse 3D occupancy decoder. Our approach distinguishes between foreground and background within a scene. Initially, foreground objects are detected using the DETR-like object detection. Subsequently, queries for both foreground and background objects are fed into the mixed dense-sparse 3D occupancy decoder, performing upsampling in dense and sparse methods, respectively. Finally, a MaskFormer is utilized to infer the semantics of the background voxels. Our approach strikes a balance between efficiency and accuracy, achieving faster inference times, lower resource consumption, and improved performance for small object detection. We demonstrate the effectiveness of our proposed method on the SemanticKITTI dataset, showcasing an mIoU of 14 and a processing speed of 10 FPS, thereby presenting a promising solution for real-time 3D semantic occupancy perception.

\end{abstract}

\section{INTRODUCTION}

3D semantic perception serves as a fundamental capability for robotics. The prevalent approach currently employs multi-sensor fusion involving LiDAR and camera \cite{uber,pointfusion,MVX-Net}, which has achieved strong performance and resulted in numerous product implementations. However, this approach raises concerns such as high costs and lack of portability. In recent years, there has been burgeoning interest in visual-based 3D semantic perception schemes due to their relatively lower cost without compromise on performance. Initially, Bird's Eye View (BEV) perception \cite{lss,bevformer,bevdet,hdmapnet,maptr} was introduced, which significantly enhanced perceptual capabilities in autonomous driving scenarios. Subsequently, 3D semantic occupancy perception \cite{monoscene,occdepth,tpvformer,voxformer,occformer} emerged, extending BEV perception by a vertical dimension, thereby offering wider applicability across various scenarios. To this end, this paper focuses on visual-based 3D semantic occupancy perception and our aim is to develop a more straightforward and more efficient method for this task.

\begin{figure} 
  \centering
  \includegraphics[width=8cm]{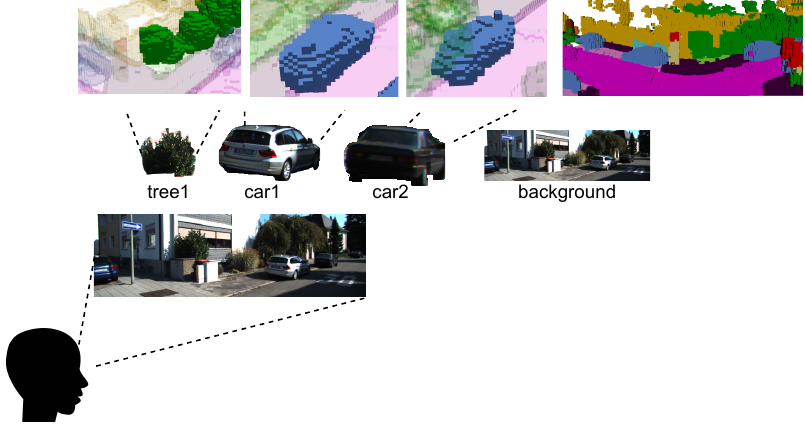}
  \caption{Concept diagram of 3D semantic occupancy perception inspired by human high-level visual processing.}
\label{fig:human_visual}
\end{figure}

\begin{figure*} 
  \centering
  \includegraphics[width=16cm]{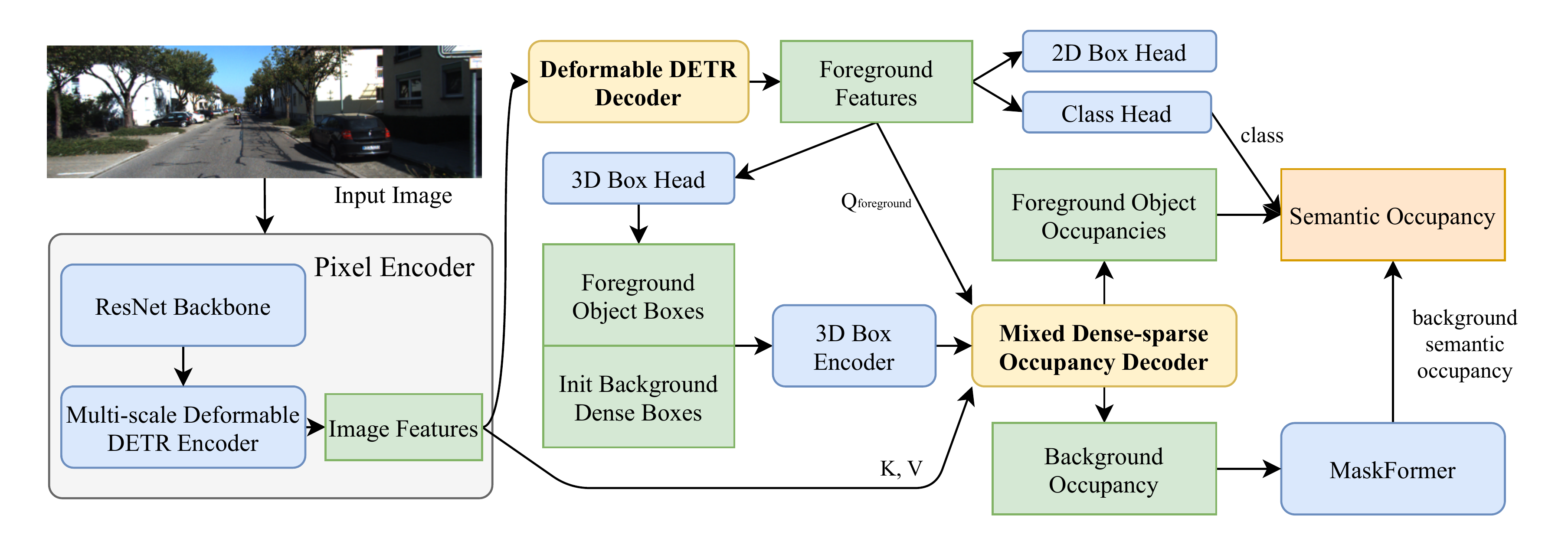}
  \caption{The system diagram of OccupancyDETR. Arrows indicate how data flow within the system.}
\label{fig:overview}
\end{figure*}

Scientists have noted in their study of human visual perception that the human brain, in high-level vision processing, tends to prioritize some objects within a scene while referencing unnoticed background information\cite{neural}. This instinct derives from humans' need to quickly identify prey and threats during wild survival. Inspired by this insight into human visual perception, we emulate this mechanism in our method for 3D semantic occupancy perception by categorizing objects within the scene into foreground and background objects, with foreground including categories such as cars and pedestrians, and background including categories like roads and trees. foreground categories possess distinctive, localized characteristics; In contrast, it is diffcult to label individual objects for background categories\textemdash for instance, labeling each tree within a forest. Moreover, background objects is heavily influenced by their spatial relationships; for instance, there exists a strong correlation between roads and buildings. 

In our work, for foreground objects, a DETR-like object detection has been introduced to guide the prediction of 3D semantic occupancy grids. The detected objects' 3D bounding box results serve as spatial priors, and their features serve as context. Together with the queries for the background objects, these are fed into a mixed dense-sparse 3D occupancy decoder. Within each decoder layer, the queries associated with foreground objects are processed through dense upsampling, while the background objects' queries are processed through sparse upsampling. Subsequently, we apply a MaskFormer\cite{maskformer} to retrieve masks for various background categories. Object detection algorithms, from early methods like YOLO\cite{yolo} to recent methods like Deformable DETR\cite{deformabledetr}, have been developed over many years, achieving excellent performance under complex scenarios. By distinguishing between foreground and background, and integrating object detection with 3D semantic occupancy perception, our method leverages these two methods. Finally, we validated our method on the SemanticKITTI dataset\cite{SemanticKITTI}, demonstrating superior performance on smaller objects, faster speed, and less resource demand.

The main contributions of this paper are as follows:
\begin{itemize}
\item We propose a novel 3D semantic occupancy prediction framework that incorporates object detection. This framework is simple and efficient, especially adept at handling small objects, and has achieved excellent performance on the SemanticKITTI dataset.
\item In response to the slow convergence problem of Detr-like algorithms, we propose a early matching pre-training. This pre-training enhances the certainty of the training and accelerates convergence.
\item To combine object detection with 3D semantic occupancy prediction, we have devised a mixed dense-sparse occupancy decoder. This decoder leverages the results of object detection as priors to enhance performance for foreground objects. And it employs dense decoding methods for foreground objects and sparse decoding for background objects, thereby improving inference efficiency and reducing resource usage.
\end{itemize}

\section{RELATED WORK}

\subsection{Camera-based BEV Perception}

In the context of autonomous driving, employing BEV as a representation of 3D space is highly suitable, especially for scenarios involving multiple cameras mounted all around the vehicle. Numerous recent studies on BEV perception has emerged, which can be mainly categorized into two methods. The first one involves LSS and its subsequent work\cite{lss, lss1, lss2, lss3}, which map 2D features to 3D space using depth estimated by a dedicated depth-estimation network, eventually obtaining BEV-level features through grid sampling. On the other hand, the second method is represented by BEVFormer\cite{bevformer}, which does not require depth estimation. Instead, it treats each cell in the 3D space as a query, establishes a mapping between that cell and 2D image features by projecting it onto the image, and then updates BEV-level features using deformable attention\cite{deformabledetr}.

\subsection{3D Semantic Occupancy Prediction}

3D semantic occupancy prediction, also known as 3D semantic scene completion (SSC), has seen some novel approaches being proposed recently. One such approach involves converting 2D features into 3D space through depth prediction, as introduced in MonoScene and OccDepth\cite{monoscene,occdepth}. Another method, similar to BEVFormer, employs a query to aggregate 2D features directly in the 3D space; examples of this approach include VoxFormer, TPVFormer SurroundOcc, etc\cite{voxformer,tpvformer,surroundocc}. MonoScene\cite{monoscene} built the first monocular method for semantic occupancy prediction, utilizing a 3D UNet to handle voxel features generated from line-of-sight projection. TPVFormer\cite{tpvformer} proposed a tri-perspective view representation to describe 3D scenes. Despite its simplicity, the tri-plane format is susceptible to a deficiency of fine-grained semantic information, leading to subpar performance. OccFormer\cite{occformer} uses a dual-path transformer network, decomposing the heavy 3D processing into local and global transformer pathways along the horizontal plane. In this paper, we propose using object detection information as a prior for 3D semantic occupancy prediction.

\subsection{Mask classification-based method}

Image semantic segmentation has traditionally been addressed as a per-pixel classification task. However, Max-Deeplab and MaskFormer approach semantic segmentation as a mask classification task\cite{MaX-DeepLab,maskformer}. This not only naturally links semantic-level segmentation with instance-level segmentation but also yields better results in semantic segmentation compared to per-pixel classification methods. Inspired by this perspective, our study models 3D occupancy prediction as a mask classification problem.

\section{APPROACH}

\subsection{Overview}
An overview of our system is shown in Fig.\ref{fig:overview} and it generally consists of two main parts\textemdash object detection module and mixed dense-sparse 3D occupancy decoder. For a input image, we initially employ the ResNet50 backbone\cite{resnet50} to extract features, followed by passing these multi-scale features into a deformable encoder for further encoding. Then, a fixed number of queries are decoded via a deformable DETR decoder as $F_{foreground}$ before being passed to three heads\textemdash class, 2D box, and 3D box heads. The results from the class and 2D box head are conventional results in object detection, with objects of high confidence being selected according to the class head's output. Regarding the 3D boxes of these objects, they are encoded by a 3D box encoder and, together with $F_{foreground}$, forms the initial queries of foreground objects. These are appended with the encoding of initial 3D boxes equally distributed in the whole space for background objects, serving as the queries for the mixed dense-sparse occupancy decoder. Passing through this decoder results in occupancy grids for each foreground and background object. The final component of the network is MaskFormer, which classifies the voxels within the grid that are occupied by background objects. Since background objects cannot be distinguished with precision, we employ semantic queries. Semantic queries, after MaskFormer, yield mask embeddings. These embeddings, when multiplied with the sparse 3D features of background objects, result in the occupancy grid for each semantic category. In the end, foreground and background objects are mapped onto the entire grid to finish the semantic occupancy perdiction.

\subsection{Object Detection Module}
In our work, we introduce object detection into 3D semantic occupancy prediction, as a means to provide priors for 3D semantic occupancy prediction. Thus, the "object" identified are different from those in conventional object detection. When generating annotated data, we first cluster semantic objects from the voxel grid based on distance, without precisely distinguishing each object. Each clustered object is then projected onto the 2D image and 2D bounding boxes is computed based on these projection points. Indeed, we consider occlusion during the projection process. We prevent completely occluded unobservable objects from affecting model learning, hence such objects are excluded. Yet, to endow the model with scene completion capabilities, partially occluded objects are preserved.

\begin{figure}[!h] 
  \centering
  \includegraphics[width=8cm]{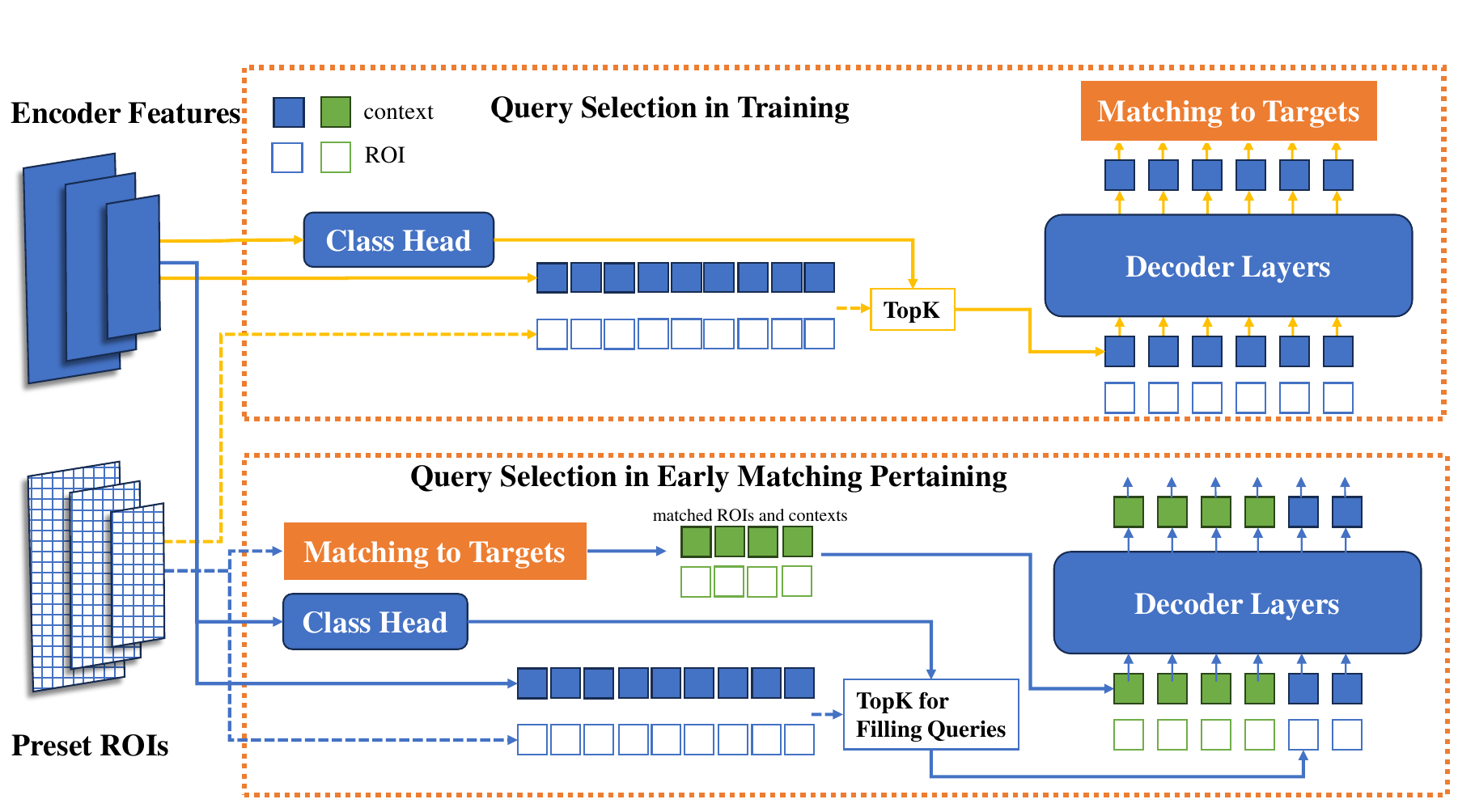}
  \caption{Data flow diagram of the object detection module, with yellow and blue arrows representing the data flows in training and early matching pretraining respectively.}
\label{fig:detr_decoder}
\end{figure}

We made improvements based on the two-stage Deformable DETR, as shown in Fig.\ref{fig:detr_decoder}. As the first end-to-end method based on transformers, DETR\cite{detr} is considered a new direction for object detection by many scholars due to its independence from any manual design. However, the unclear queries in DETR and the uncertainties brought by bipartite matching cause the convergence extremely slow during training. We found that during the prolonged training of DETR, most of the time, bipartite matching remains unstable. We attribute this to the fact that the model has to go through numerous of trials to find appropriate queries for the whole dataset, which takes up most of the training time. Hence, in the object detection module, we adopt the two-stage deformable DETR and designed an early matching pretraining for the query selection process. During the regular training phase, each multi-scale feature output from the encoder is assigned a preset ROI in the query selection process. These features are computed through a classification head, and the top k features with the highest scores are selected as the context of queries, with their corresponding ROIs acting as the position of queries. After passing through the deformable decoder, they are then matched with groundtruth. In the early matching pretraining, the preceding bipartite matching between the preset ROIs and groundtruth ensures certainty, avoiding the prolonged process of searching suitable queries, thereby accelerating the following regular training.

In the final stage of object detection, the queries processed by the deformable detr decoder already have vague 3D spatial information. Therefore, in addition to the classification head and 2D bounding box head, we add an extra 3D bounding box head. This is used to predict the 3D bounding box of the object in the camera coordinate system. Then, according to the camera extrinsic, it is transformed into the occupancy grid coordinate system to provide position priors for the subsequent mixed dense-sparse 3D occupancy decoder.

\begin{table*}
  \centering
  \vspace{10pt}
  \caption{Semantic scene completion result on SemanticKITTI test set.}
  \label{Table1}
  \resizebox{\linewidth}{!}{
  \begin{tabular}{c|ccccccccccccccccccccc} 
  \toprule
  \multirow{2}{*}{Method} & \multicolumn{20}{c}{SSC} \\
  \cline{2-21} & 
  \rotatebox{90}{\tikz \fill[car] (0,0) rectangle (0.2,0.2); car (3.92\%)} & 
  \rotatebox{90}{\tikz \fill[bicycle] (0,0) rectangle (0.2,0.2); bicycle (0.03\%)} & 
  \rotatebox{90}{\tikz \fill[motorcycle] (0,0) rectangle (0.2,0.2); motorcycle (0.05\%)} & 
  \rotatebox{90}{\tikz \fill[truck] (0,0) rectangle (0.2,0.2); truck (0.16\%)} & 
  \rotatebox{90}{\tikz \fill[other-vehicle] (0,0) rectangle (0.2,0.2); other-vehicle (0.20\%)} &  
  \rotatebox{90}{\tikz \fill[person] (0,0) rectangle (0.2,0.2); person (0.07\%)} &  
  \rotatebox{90}{\tikz \fill[bicyclist] (0,0) rectangle (0.2,0.2); bicyclist (0.07\%)} &  
  \rotatebox{90}{\tikz \fill[motorcyclist] (0,0) rectangle (0.2,0.2); motorcyclist (0.05\%)} &  
  \rotatebox{90}{\tikz \fill[road] (0,0) rectangle (0.2,0.2); road (15.30\%)} &  
  \rotatebox{90}{\tikz \fill[parking] (0,0) rectangle (0.2,0.2); parking (11.12\%)} &  
  \rotatebox{90}{\tikz \fill[sidewalk] (0,0) rectangle (0.2,0.2); sidewalk (11.13\%)} &  
  \rotatebox{90}{\tikz \fill[other-ground] (0,0) rectangle (0.2,0.2); other-ground (0.56\%)} &  
  \rotatebox{90}{\tikz \fill[building] (0,0) rectangle (0.2,0.2); building (14.1\%)} &  
  \rotatebox{90}{\tikz \fill[fence] (0,0) rectangle (0.2,0.2); fence (3.90\%)} &  
  \rotatebox{90}{\tikz \fill[vegetation] (0,0) rectangle (0.2,0.2); vegetation (39.3\%)} &  
  \rotatebox{90}{\tikz \fill[trunk] (0,0) rectangle (0.2,0.2); trunk (0.16\%)} &  
  \rotatebox{90}{\tikz \fill[terrain] (0,0) rectangle (0.2,0.2); terrain (9.17\%)} &  
  \rotatebox{90}{\tikz \fill[pole] (0,0) rectangle (0.2,0.2); pole (0.29\%)}  & 
  \rotatebox{90}{\tikz \fill[traffic-sign] (0,0) rectangle (0.2,0.2); traffic-sign (0.08\%)} & 
  mIoU \\
  \midrule
  MonoScene & 23.26 & 0.61 & 0.45 & 6.98 & 1.48 & 1.86 & 1.20 & 0.0 & 56.52 & 14.27 & 26.72 & 0.46 & 14.09 & 5.84 & 17.89 & 2.81 & 29.64 & 4.14 & 2.25 & 11.08\\
  TPVFormer & 23.81 & 0.36 & 0.05 & 8.08 & 4.35 & 0.51 & 0.89 & 0.0 & 56.5 & 20.6 & 25.87 & \textbf{0.85} & 13.88 & \textbf{5.94} & 16.92 & 2.26 & 30.38 & 3.14 & 1.52 & 11.36 \\
  OccFormer & 25.09 & 0.81 & 1.19 & \textbf{25.53} & 8.52 & 2.78 & 2.82 & 0.0 & \textbf{58.85} & \textbf{19.61} & \textbf{26.88} & 0.31 & \textbf{14.4} & 5.61 & \textbf{19.63} & \textbf{3.93} & \textbf{32.62} & 4.26 & 2.86 & 13.46 \\
  OccupancyDETR & \textbf{25.41} & \textbf{5.41} & \textbf{4.56} & 24.42 & \textbf{11.91} & \textbf{6.31} & \textbf{5.98} & 0.0 & 56.54 & 16.21 & 26.34 & 0.53 & 14.36 & 5.84 & 17.82 & 3.12 & 30.81 & \textbf{7.19} & \textbf{5.13} & \textbf{14.10} \\
  \midrule
  OccupancyDETR(all foreground) & 25.32 & 5.43 & 4.72 & 24.14 & 11.85 & 6.43 & 6.03 & 0.0 & 36.68 & 11.24 & 15.67 & 0.40 & 9.92 & 4.03 & 10.23 & 3.06 & 18.61 & 7.13 & 5.39 & 10.86 \\
  OccupancyDETR(all background) & 23.12 & 0.64 & 0.54 & 7.13 & 4.45 & 2.21 & 1.50 & 0.0 & 56.55 & 16.11 & 26.41 & 0.51 & 14.24 & 5.81 & 17.74 & 3.12 & 30.72 & 4.10 & 2.15 & 11.42 \\
  \bottomrule
  \end{tabular}}
  \label{tab:moiu}
\end{table*}

\subsection{Mixed Dense-sparse 3D Occupancy Decoder}

\begin{figure*}[!h] 
  \centering
  \includegraphics[width=16cm]{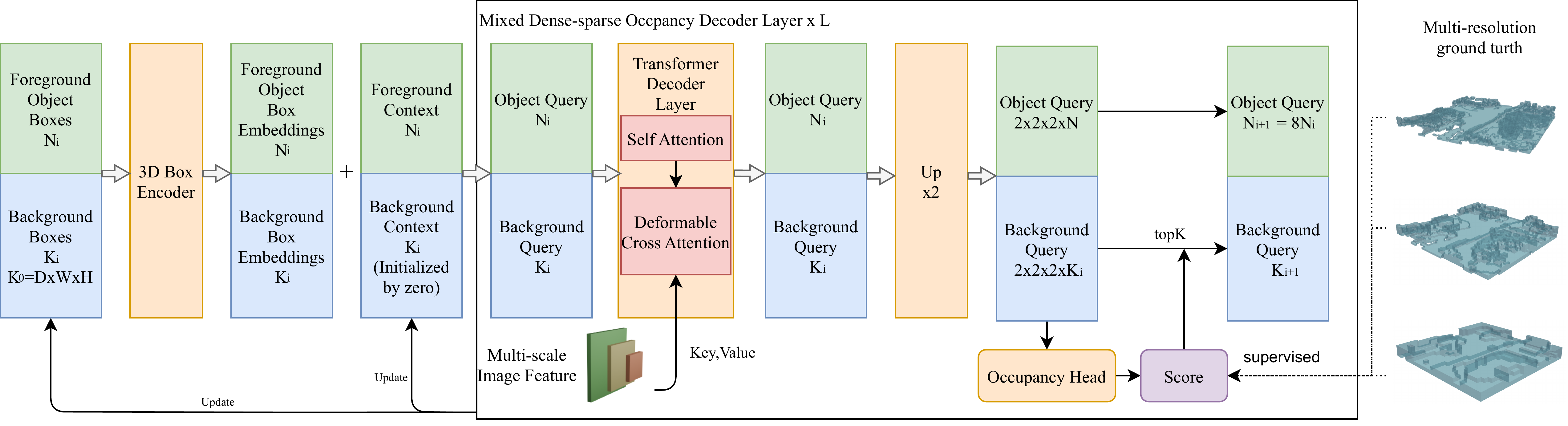}
  \caption{The detailed architecture of mixed dense-sparse occupancy decoder}
\label{fig:3d_occ_decoder}
\end{figure*}

The mixed dense-sparse occupancy decoder employs a strategy wherein dense occupancy decoders are utilized for foreground objects, while sparse occupancy decoders akin to SparseOcc\cite{sparseocc} are employed for background objects. This strategy ensures that the full extent of information is preserved during the occupancy decoding for foreground objects, while simultaneously avoiding computations on the unoccupied voxels, which constitute approximately 90\% of the entire grid, during background occupancy decoding. This results in a reduction of resource utilization.

In the case of foreground objects, object detection achieves notable results on the 2D level; however, due to a deficiency in feature extraction on the 3D level, the precision of the 3D bounding box predictions is compromised. Accordingly, the reliability of these 3D bounding boxes predictions cannot be wholly assured. Because of this, the function of the 3D bounding boxes is only to furnish priors, while high-resolution occupancy grids is infered by the mixed dense-sparse occupancy decoder. Therefore, we enlarge the length, width, and height of all 3D bounding boxes by 50\% to ensure that they can encapsulate the entire object.

As depicted in Fig.\ref{fig:3d_occ_decoder}, the mixed dense-sparse 3D occupancy decoder is composed of $L$ layers of occupancy decoder layers. Each layer sequentially performs self-attention among the 3D queries followed by deformable cross-attention with multi-scale image features. Let the initial shape of queries for background objects be $(D_0, W_0, H_0)$, and the number of foreground objects is $N_0$. Upsampling is performed every layer, where the output is upsampled by $\times 2$, and the 3D query boxes are subdivided into 8 smaller 3D boxes arranged in $(2, 2, 2)$. After upsampling, all foreground object 3D query boxes are preserved, while for background objects, only the top-K 3D query boxes are retained based on their scores obtained from the occupancy head (where K is a parameter set according to the dataset and resolution). Ground truth occupancy grids at various resolutions are precomputed for supervision of the occupancy head. After every layers, both the quantity of queries and the size of the 3D boxes are changed, constituting a coarse-to-fine process. 

We have obtained dense voxel features of the foreground objects and the K most likely occupied voxel features of the background objects. Subsequently, MaskFormer is employed to classify the voxels of the background objects. Based on the classification results, the foreground and background objects are mapped into the grid as 3D boxes to yield the final semantic occupancy grid.

\begin{figure*}[h]
  \begin{minipage}[t]{1\linewidth}
  \centering
  \includegraphics[width=16cm]{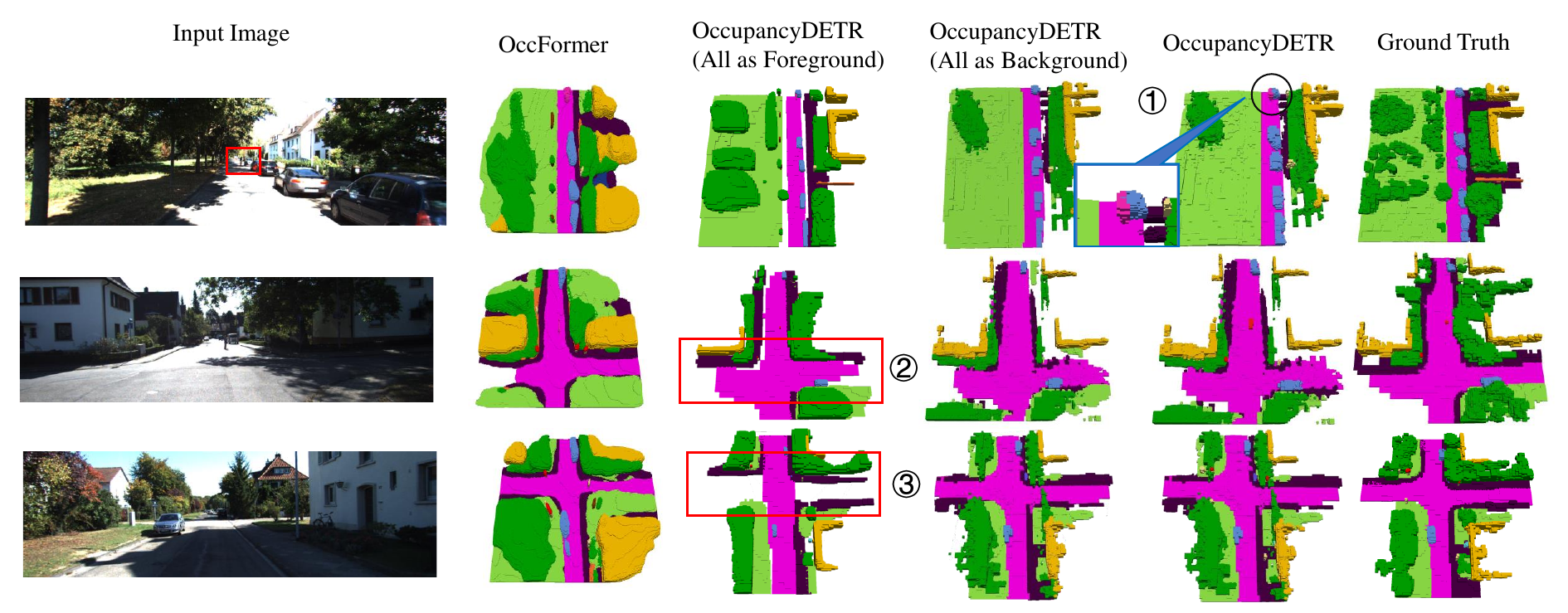}
  \end{minipage}%

  \begin{minipage}[t]{1\linewidth}
    \centering
    \footnotesize{
    \tikz \fill[car] (0,0) rectangle (0.2,0.2); car
    \tikz \fill[bicycle] (0,0) rectangle (0.2,0.2); bicycle
    \tikz \fill[motorcycle] (0,0) rectangle (0.2,0.2); motorcycle
    \tikz \fill[truck] (0,0) rectangle (0.2,0.2); truck
    \tikz \fill[other-vehicle] (0,0) rectangle (0.2,0.2); other-vehicle
    \tikz \fill[person] (0,0) rectangle (0.2,0.2); person
    \tikz \fill[bicyclist] (0,0) rectangle (0.2,0.2); bicyclist
    \tikz \fill[motorcyclist] (0,0) rectangle (0.2,0.2); motorcyclist
    \tikz \fill[road] (0,0) rectangle (0.2,0.2); road
    \tikz \fill[parking] (0,0) rectangle (0.2,0.2); parking}

    \footnotesize{
    \tikz \fill[sidewalk] (0,0) rectangle (0.2,0.2); sidewalk
    \tikz \fill[other-ground] (0,0) rectangle (0.2,0.2); other-ground
    \tikz \fill[building] (0,0) rectangle (0.2,0.2); building
    \tikz \fill[fence] (0,0) rectangle (0.2,0.2); fence
    \tikz \fill[vegetation] (0,0) rectangle (0.2,0.2); vegetation
    \tikz \fill[trunk] (0,0) rectangle (0.2,0.2); trunk
    \tikz \fill[terrain] (0,0) rectangle (0.2,0.2); terrain
    \tikz \fill[pole] (0,0) rectangle (0.2,0.2); pole
    \tikz \fill[traffic-sign] (0,0) rectangle (0.2,0.2); traffic-sign}
  \end{minipage}%
  \caption{Qualitative results on SemanticKITTI test set.}

\label{fig:experiments}
\end{figure*}

\subsection{Training strategy}

The entire training process is divided into two steps. The first step involves object detection module pre-training by early matching. This step is designed to expedite the convergence of the two-stage deformable DETR. The second step encompasses end-to-end training of the entire network.

\textbf{Loss Function} Below is the process for computing the loss associated with 3D occupancy. We extract local ground turth occupancy grids for each foreground object according to its predicted 3D bounding box, which are then scaled to $(2^L, 2^L, 2^L)$ through linear interpolation. The background ground truth grid is interpolated and scaled to $(D_0 2^L, W_0 2^L, H_0 2^L)$, and subsequently downsampled L times to match the sparse occupancy grid results at the (L)th level of the mixed dense-sparse occupancy decoder, as shown on the right side of Fig.\ref{fig:3d_occ_decoder}. The supervisory loss for foreground occupancy is determined using the dice loss\cite{diceloss}, while the background occupancy loss employs the binary cross-entropy (BCE) loss. For object detection, the classification loss is calculated using the focal loss, and the loss for both 2D and 3D boxes is a composite of L1 loss and generalized intersection over union (GIoU) loss. MaskFormer utilizes a combination of focal loss and dice loss. The overall loss computation is as follows:


\begin{equation}
\begin{aligned}
\mathcal{L} = &\mathcal{L}_{cls}+\lambda_{\text {box2d }}\mathcal{L}_{\text {box2d }}+\lambda_{\text {box2d }}\mathcal{L}_{\text {giou2d }} &\\
& +\lambda_{\text {box3d }}\mathcal{L}_{\text {box3d }} +\lambda_{\text {giou3d }}\mathcal{L}_{\text {giou3d }} &\\
& + \lambda_{\text {background }}\mathcal{L}_{\text {BCE }} + \mathcal{L}_{\text {maskformer }}  &\\
& + \lambda_{\text {foreground }}\mathcal{L}_{\text {dice }} &\\
\mathcal{L}_{\text {maskformer }} = &\lambda_{\text {focal }}\mathcal{L}_{\text {focal }} + \lambda_{\text {dice2 }}\mathcal{L}_{\text {dice2 }}
\label{eq:my_equation}
\end{aligned}
\end{equation}

\section{EXPERIMENTS}

\subsection{Datasets}

\begin{figure}[h]
  \centering
  \subfigure[2D labels]{
  \begin{minipage}[t]{0.9\linewidth}
  \includegraphics[width=0.9\linewidth]{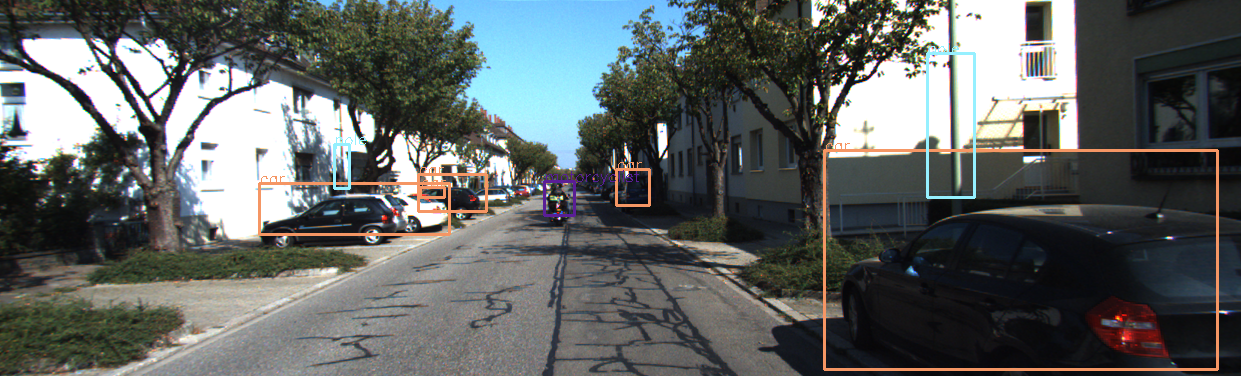}
  \end{minipage}%
  }%

  \subfigure[3D labels]{
  \begin{minipage}[t]{0.9\linewidth}
  \centering
  \includegraphics[width=0.9\linewidth]{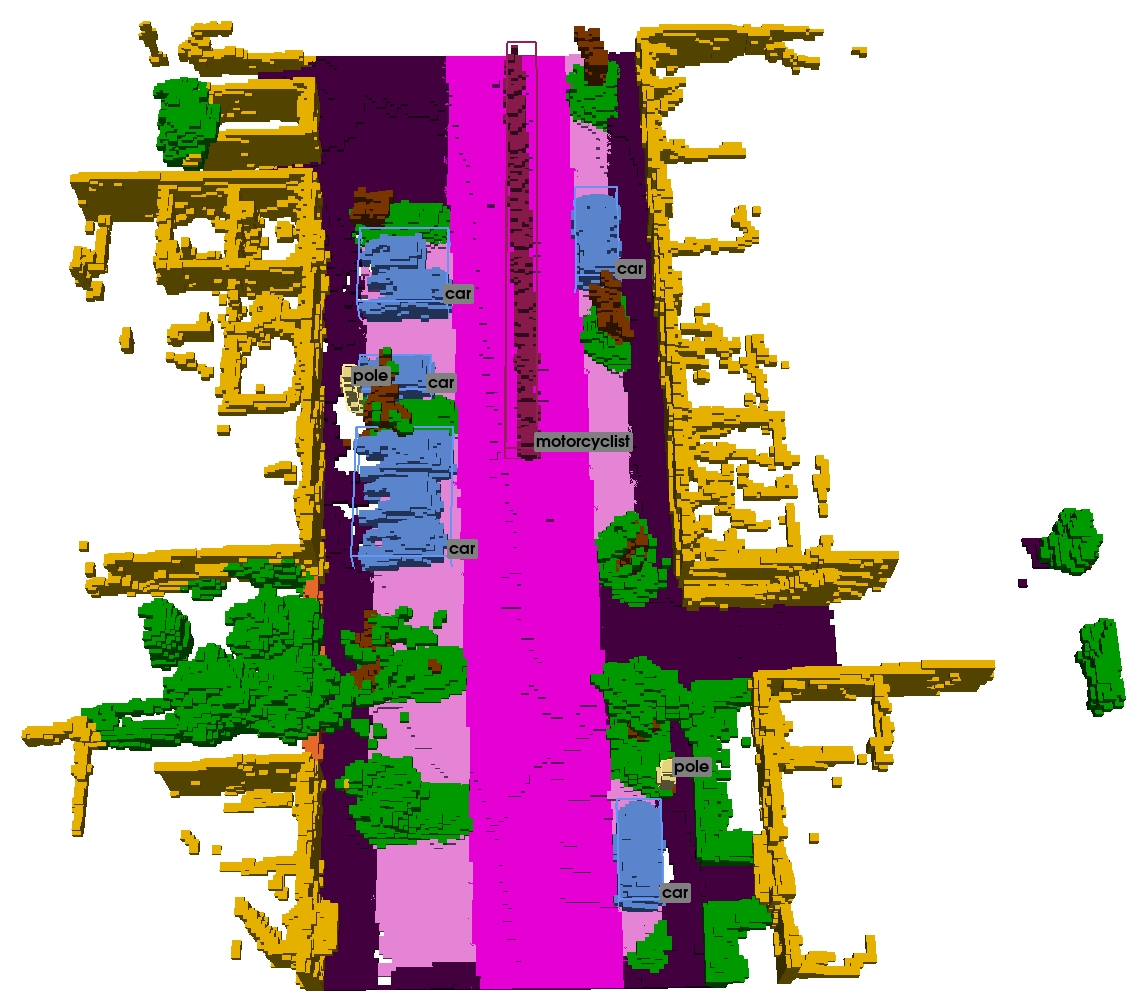}
  \end{minipage}%
  }%
  \centering
  \caption{Examples of 2D and 3D labels obtained after object extraction.}

\label{fig:semantickitti}
\end{figure}

The SemanticKITTI dataset\cite{SemanticKITTI}, built upon the widely popular KITTI Odometry dataset\cite{kitti}, emphasizes on semantic understanding of scenes using LiDAR points and forward-facing cameras. OccupancyDETR, functioning as a monocular 3D semantic occupancy perception, employs only the left front-view camera as input. Within this dataset, the annotated semantic occupancy is represented as a voxel grid with a shape of $256\times 256\times 32$. Each voxel measures $0.2m\times 0.2m\times 0.2m$ and carries labels for 21 semantic categories (19 semantics, 1 free, 1 unknown). We designate \textit{car}, \textit{bicycle}, \textit{motorcycle}, \textit{truck}, \textit{other-vehicle}, \textit{person}, \textit{bicyclist}, \textit{motorcyclist}, \textit{pole}, and \textit{traffic-sign} as foreground categories, and \textit{road}, \textit{parking}, \textit{sidewalk}, \textit{other-ground}, \textit{building}, \textit{fence}, \textit{vegetation}, \textit{trunk}, and \textit{terrain} as background categories. The 2D and 3D labels of the dataset obtained after object extraction are shown in Fig.\ref{fig:semantickitti}.

\subsection{Experiments Setup}

The model training is conducted on a Nvidia RTX 3090 GPU (24G). The training process span two stages, with respective epochs of 50 epochs, 50 epochs. The initial learning rates for each stage are set as 1e-4, 1e-4, and then linearly decrease to zero. AdamW with weight decay 0.01 is employed as the optimizer. We configure the encoder for object detection with 6 layers, the decoder with 6 layers, the mixed dense-sparse occupancy decoder with 3 layers, and MaskFormer with 4 layers.

\subsection{Main Results}

As shown in Table \ref{tab:moiu}, we report the mean intersection over union (mIoU) for semantic scene completion (SSC) task. We compare it against other monocular 3D semantic occupancy perception methods and two modes of our method: one where all categories are treated as foreground and another where all categories are treated as background. Representative cases are displayed in Fig.\ref{fig:experiments}. At \ding{172}, we demonstrate the performance of our method on smaller object categories. Our approach can detect distant bicyclists, which may not be identified under the "all categories as background" mode due to the sparse upsampling that potentially discards smaller distant objects. For the "all categories as foreground" mode, \ding{173} and \ding{174} reveal deficiencies in performance on the \textit{road} and \textit{sidewalk} categories. We believe that these categories are not well-suited for object detection because object detection have limited capability in extracting features that regard the relationships between different objects in 3D space. This limitation results in this mode being able to directly detect nearby crossroads, but failing to indirectly infer more distant crossroads based on the nearby buildings. Overall, our method achieves commendable performance on both foreground and background objects.

\begin{table}
  \centering
  \vspace{10pt}
  \caption{Model parameters and single image inferrence time on SemanticKITTI test set.}
  \label{Table2}
  \resizebox{\linewidth}{!}{
  \begin{tabular}{ccccc}
  \toprule
  Method & Parameters & Inferrence Time \\
  \midrule
  MonoScene & 149.55M & 646ms \\
  TPVFormer & 140.55M & 424ms \\
  OccFormer & 203.68M & 340ms \\
  OccupancyDETR   & 51.11M & \textbf{95ms} \\
  \bottomrule
  \end{tabular}}
  \label{tab:time}
\end{table}

Moreover, the advantage of our method in terms of speed and resource requirements is significant. The inference time pre frame and parameter count of our method are shown in Table \ref{tab:time}. Our method has an average inference time of 95ms, which is close to real-time performance. Inferrence time is measured on a Nvidia RTX 3090 GPU (24G) with the PyTorch fp32 backend (single batch size).

\subsection{Comparative Study on Early Matching Pretraining}

\begin{figure}[!h] 
  \centering
  \includegraphics[width=8cm]{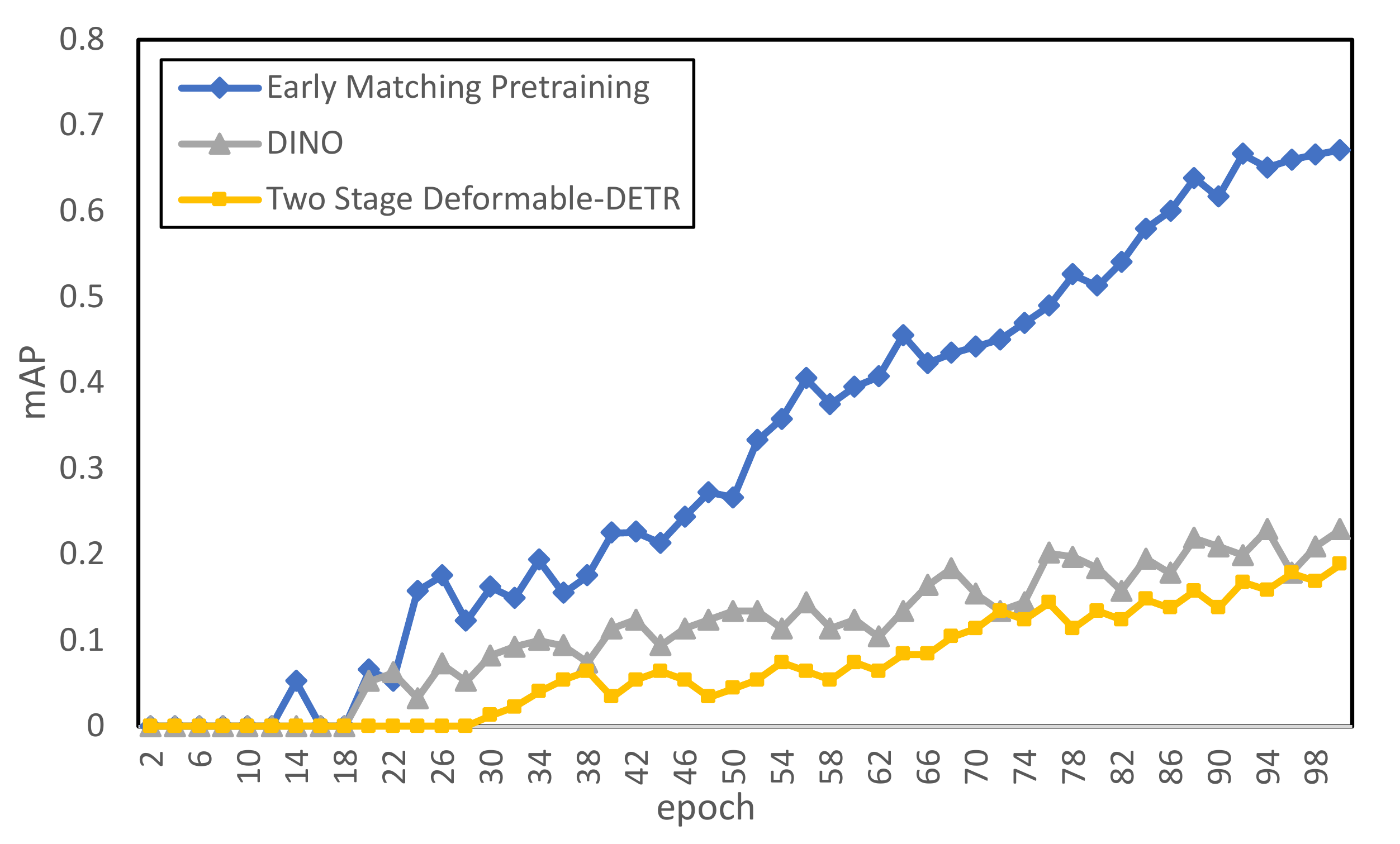}
  \caption{Training convergence curves evaluated on SemanticKITTI for early matching pretraining, DINO and two stage Deformable DETR.}
\label{fig:earlymatching}
\end{figure}

In order to validate the positive impact of early matching pretraining on the training of DETR-like \cite{detr} object detection models, we conduct a comparative study under the same experimental conditions between DINO\cite{dino} and two stage Deformable DETR\cite{deformabledetr}. In our experiment, we set the initial learning rate at 1e-4 and decrease it linearly to zero within 100 epochs. Fig.\ref{fig:earlymatching} illustrates the mAP curves for the three methods on the test set during training, showing that early matching pretraining leads to faster convergence in training. Furthermore, we analyze DINO's performance, which is a method based on the two stage Deformable DETR. Several improvements are proposed by DINO to accelerate convergence, one of which is mixed query selection. This process involves using learnable embeddings as static content queries while selecting anchors through query selection to serve as dynamic anchors. However, a misalignment issue exists between the orders of static content queries and dynamic anchors. We hypothesize that this discrepancy is the reason DINO's performance does not meet expectations.

\section{CONCLUSIONS}

In this paper, we propose a method of 3D semantic occupancy perception combined with object detection, using a mixed dense-sparse occupancy decoder. Our method not only reduces the required resources and accelerates inference speed but also enhances performance on small objects. Furthermore, to address the slow training issue associated with DETR, we propose a early matching pretraining strategy. Experiments on the SemanticKITTI dataset have demonstrated commendable results, affirming the simplicity and efficiency of our method. Additionally, our framework is designed to be easily adaptable to multi-frame and multi-camera input, an avenue we aim to pursue in our future research endeavors.




\bibliographystyle{IEEEtran}  
\bibliography{references}

\end{document}